\documentclass{article}

\usepackage{PRIMEarxiv}

\usepackage[utf8]{inputenc}
\usepackage[T1]{fontenc}
\usepackage{amsmath}
\usepackage{amsfonts}
\usepackage{amssymb}
\usepackage{graphicx}
\usepackage{booktabs}
\usepackage{multirow}
\usepackage{xspace}
\usepackage[table]{xcolor}
\usepackage{algorithm}
\usepackage{algorithmic}
\usepackage{caption}
\usepackage{microtype}
\usepackage[hyphens]{url}
\usepackage{hyperref}

\providecommand{\method}{\textsc{SIMPLE}\xspace}
\title{Pretraining Reusable Inference Across Views with Synthetic Task Priors}

\author{
\textbf{Jielong Lu$^{1}$ \quad Zhihao Wu$^{1}$ \quad Jiajun Yu$^{1}$} \quad 
\textbf{Zhaoliang Chen$^{2}$ \quad Haishuai Wang$^{1}$} \\[2pt]
{\normalfont $^{1}$Zhejiang University, Hangzhou, China} \\
{\normalfont $^{2}$Hong Kong Baptist University, Hong Kong SAR, China}
}

\begin{document}

\maketitle

\begin{abstract}
Modern pretrained encoders make representations from heterogeneous views increasingly reusable, but the procedure that determines view utility and combines evidence is still relearned for each downstream task. Consequently, knowledge about view relevance, complementarity, reliability, and missingness is repeatedly discarded rather than transferred across tasks. We therefore reformulate multi-view learning as learning a reusable, task-conditioned inference procedure rather than a fixed fusion function.
Based on this perspective, we propose \method, a prior-fitted multi-view in-context learner that predicts query labels by conditioning on a small labeled support set. Since existing real-world datasets cover only a limited range of view configurations and task structures, we construct a controllable synthetic task prior in embedding space. It generates diverse support-query episodes with varying class structures, shared and view-specific factors, representation geometries, cross-view dependencies, reliability levels, missingness patterns, and distribution shifts. A hierarchical inference architecture then performs reasoning within views, across views, and across support and query samples.
Experiments on multi-view and multi-omics benchmarks demonstrate that the frozen variant of \method achieves competitive performance without updating the inference backbone, while lightweight adapter calibration attains leading performance on most evaluated datasets. Together, the results under frozen, one-shot, and missing-view settings support the central hypothesis that multi-view reasoning itself can be pretrained and reused, while lightweight adapter calibration provides task-specific alignment when needed.
\end{abstract}

\keywords{Multi-view Learning \and In-Context Learning \and Synthetic Task Prior \and Transfer Learning}

\section{Introduction}
\label{sec:introduction}

Multi-view learning integrates multiple observations of the same entity
to improve prediction \cite{yeh2026seeing,wang2026wasserstein,liu2026sparsemvc}. With the development of pretrained encoders,
representations from images, texts, graphs, sensors, and biological
measurements can increasingly be reused across downstream tasks \cite{radford2021learning,li2022blip,alayrac2022flamingo,girdhar2023imagebind}.
However, the subsequent multi-view inference stage remains largely
dataset-specific. For each new dataset, existing methods typically train a new fusion module and prediction head to determine which views are useful, how their information should be combined, and how the fused representation should be mapped to the target labels \cite{zadeh2018memory,wang2023metaviewer,hu2025adaptive,luo2026estim,tang2026trusted}.

This creates a fundamental asymmetry in modern multi-view systems:
view-specific representations are reusable, whereas the ability to reason
over views is repeatedly discarded and relearned. Knowledge about view
relevance, complementarity, redundancy, conflict, and missing-view
handling is encoded in task-specific parameters and cannot be readily
transferred to a new problem \cite{ECMGD,xu2024reliable}. 
As a result, each downstream dataset is
treated as an independent optimization problem, even though many of the
underlying inference operations are shared.

A more expressive fusion architecture alone does not resolve this issue.
The utility of a view is not fixed, but depends on the current prediction
target, the observed instance, the quality of its representation, and
which other views are available. A view may be informative for one task
but irrelevant for another, complementary to one view but redundant with
another, or become essential when alternative observations are missing.
Therefore, an appropriate fusion strategy should be inferred from the
current task rather than permanently encoded in dataset-specific
parameters.

We consequently consider a different formulation of multi-view learning.
Instead of learning one fusion function for one dataset, a model should
learn how to construct a suitable prediction rule for a new task.
Given a labeled support set and unlabeled query samples, the model should
infer the label semantics, identify task-relevant information, estimate
the conditional reliability of different views, and determine how their
evidence should be combined. Under this formulation, the support set acts
as a dataset-level task specification rather than merely as supervision
for fitting another classifier. This leads to our central question:

\begin{quote}
\emph{Can the process of solving multi-view learning problems be
pretrained once and reused across previously unseen tasks?}
\end{quote}

In \method, the support set specifies the downstream task, while a pretrained
inference backbone constructs the corresponding prediction rule through its
forward computation. The object of pretraining is therefore not a fixed
predictor for any one dataset, but a procedure for reasoning across views.
We instantiate this idea through prior-fitted in-context learning
\cite{hollmann2025tabpfn}, with the goal of amortizing task-specific
multi-view learning into a shared inference model. Training such a model, however, requires
exposure to a broad distribution of multi-view learning problems.
Existing real-world datasets provide insufficient coverage: each dataset
typically fixes the view composition, encoder representations, label
space, cross-view relationships, and observation process. It provides
many samples from one task, but only one realization of the much broader
multi-view task space.

To obtain the required task diversity, we construct a controllable
synthetic multi-view task prior directly in embedding space. Instead of
generating realistic raw views, the prior generates complete
support-query episodes that emulate the heterogeneous representations
produced by different encoders. Across episodes, we vary class structure,
shared and view-specific information, representation geometry,
cross-view redundancy and complementarity, view quality, missingness
mechanisms, and support-query distribution shifts. Importantly, the
prior generates different learning problems rather than additional
samples from a fixed problem. The model must therefore infer a new
prediction rule from context in every episode.

Based on this task prior, we propose a method named  \underline{S}ynthetic-prior \underline{I}n-context \underline{M}ulti-view \underline{P}retrained \underline{L}earn\underline{E}r (SIMPLE).
 The architecture follows the structure of
the required inference process. Feature-level reasoning identifies
task-relevant information within each view; view-level reasoning models
conditional relationships among the available views; and sample-level
reasoning transfers supervision from support examples to query instances.
This hierarchy is not an arbitrary combination of fusion modules, but an
explicit decomposition of the computations required to solve an unseen
multi-view task.
At deployment, \method supports two complementary protocols.
\method-F directly applies the pretrained inference backbone without
updating its parameters. \method-A optimizes only
lightweight view adapters and the output head while keeping the
inference backbone frozen.
Experiments on multi-view
and multi-omics benchmarks show that \method-F can remain competitive
with fully trained task-specific methods, while \method-A achieves
leading performance on most evaluated datasets. Further results under
one-shot supervision and missing-view conditions demonstrate that the
model learns a transferable multi-view inference procedure rather than a
fixed dataset-specific fusion rule.

Our main contributions are summarized as follows:

\begin{itemize}
    \item We reformulate multi-view learning as learning a reusable,
    task-conditioned inference procedure, rather than fitting an
    independent fusion function for every downstream dataset.

    \item We introduce a controllable synthetic multi-view task prior that
    generates diverse support--query learning problems with varying task
    structures, representation geometries, cross-view relationships,
    reliability levels, missingness patterns, and distribution shifts.

    \item We develop a hierarchical in-context architecture that performs
    reasoning within views, across views, and across support and query
    samples, supporting both direct frozen inference and lightweight
    adapter-based calibration.

    \item Experiments separate direct reuse from task-specific adaptation:
    \method-F evaluates frozen contextual inference, \method-A evaluates
    lightweight calibration, and the one-shot and missing-view studies probe
    limited task evidence and changing view availability. A complementary
    cross-domain study evaluates frozen-backbone transfer under domain shift.
\end{itemize}

\begin{figure*}[!t]
\centering
\includegraphics[width=\textwidth]{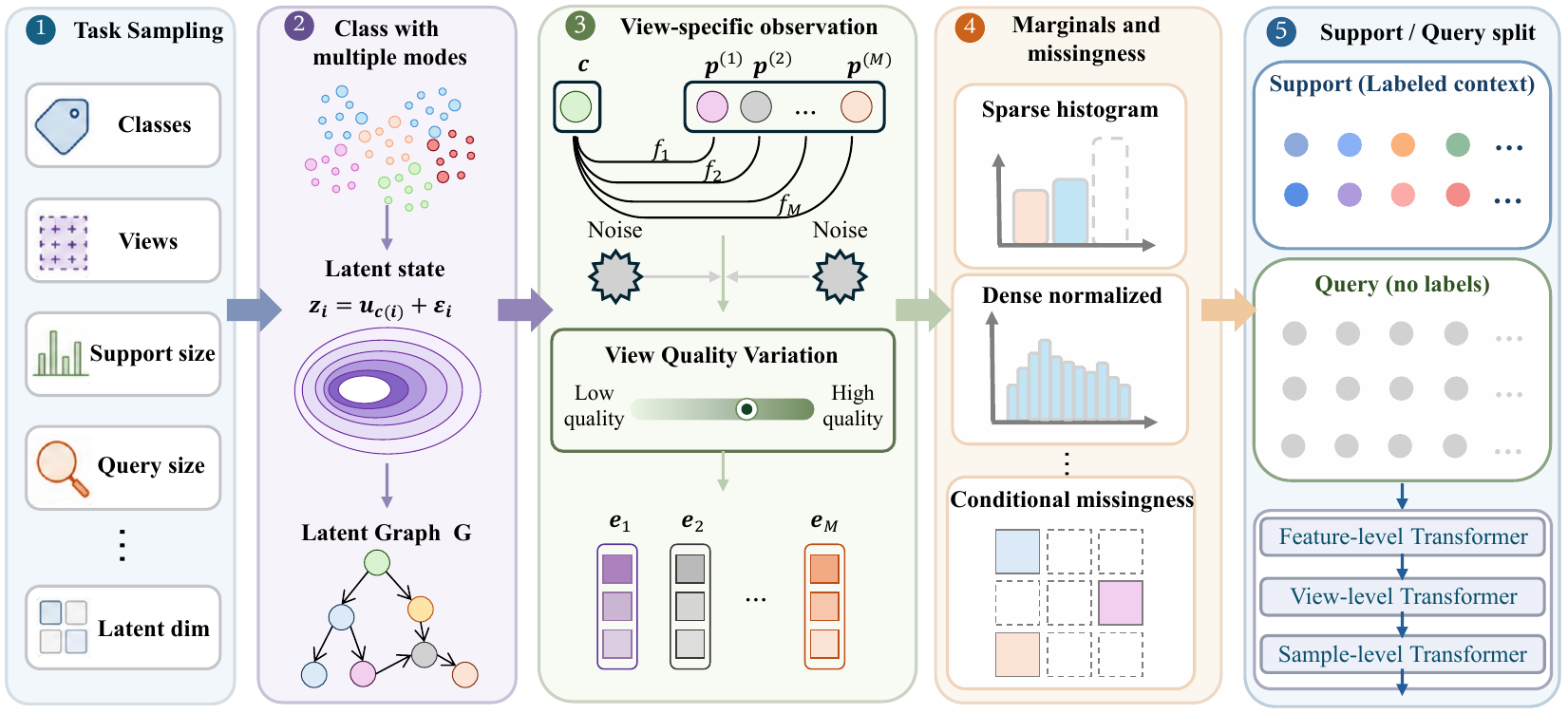}\\
\caption{The pipeline of the proposed framework SIMPLE.}
\label{Framework}
\end{figure*}

\section{Method}

\label{sec:method}

\subsection{Notation and Problem Formulation}
\label{subsec:problem_formulation}

We consider a multi-view supervised learning task containing $N$ samples and
$M$ heterogeneous views. The $i$-th sample is represented as
    $\mathcal{X}_{i}
    =
    \{
        \mathbf{x}_{i}^{(m)}
    \}_{m=1}^{M},
    \quad
    i\in\{1,\ldots,N\},$
where $\mathbf{x}_{i}^{(m)}$ denotes the raw observation of view $m$.
Depending on the application, $\mathbf{x}_{i}^{(m)}$ may represent an image,
a text sequence, an audio signal, a time series, a graph, or a structured
feature vector.
Each view is processed by a view-specific encoder
$E_m(\cdot)$:
\begin{equation}
    \mathbf{h}_{i}^{(m)}
    =
    E_m
    (
        \mathbf{x}_{i}^{(m)}
    ),
    \qquad
    \mathbf{h}_{i}^{(m)}
    \in
    \mathbb{R}^{d_m},
    \label{eq:modality_encoder}
\end{equation}
where $d_m$ denotes the output dimension of encoder $E_m$.
The view encoders may have different architectures, pretraining
objectives, and output dimensions.
A view-specific adapter $P_m(\cdot)$ maps each encoder output into a
common representation space:
\begin{equation}
    \mathbf{z}_{i}^{(m)}
    =
    P_m
    (
        \mathbf{h}_{i}^{(m)}
    ),
    \qquad
    \mathbf{z}_{i}^{(m)}
    \in
    \mathbb{R}^{d_z},
    \label{eq:modality_projection}
\end{equation}
where $d_z$ is the unified embedding dimension.

Let
\begin{equation}
    \mathcal{S}
    =
    \left\{
        (
            \{
                \mathbf{z}_{i}^{(m)}
            \}_{m=1}^{M},
            y_i
        )
    \right\}_{i=1}^{N_s}
    \label{eq:support_set}
\end{equation}
denote a labeled support set, and let
\begin{equation}
    \mathcal{Q}
    =
    \left\{
        \{
            \mathbf{z}_{j}^{(m)}
        \}_{m=1}^{M}
    \right\}_{j=N_s+1}^{N_s+N_q}
    \label{eq:query_set}
\end{equation}
denote an unlabeled query set, where $N=N_s+N_q$.
Our objective is to learn a reusable multi-view inference model
$F_{\boldsymbol{\theta}}$ such that
\begin{equation}
    F_{\boldsymbol{\theta}}
    :
    \left(
        \mathcal{S},
        \mathcal{Q}
    \right)
    \mapsto
    \widehat{\mathbf{Y}}_{\mathcal{Q}},
    \label{eq:foundation_inference_function}
\end{equation}
where
    $\widehat{\mathbf{Y}}_{\mathcal{Q}}
    =
    \left[
        \widehat{y}_{N_s+1},
        \ldots,
        \widehat{y}_{N_s+N_q}
    \right]^{\top}$
contains the predictions for all query samples.

Unlike conventional multi-view models that optimize a task-specific fusion
module for each downstream dataset, the proposed model is pretrained over a
distribution of synthetic multi-view tasks and is subsequently reused as a
general-purpose inference backbone.

\subsection{Multi-view Synthetic Task Prior}
\label{subsec:prior}

A pretrained multi-view learner requires exposure to a broad spectrum of
learning problems in order to acquire transferable inference capabilities.
However, existing multi-view datasets only represent a small subset of
possible combinations of views, semantic structures, and observation
conditions. Therefore, we construct a synthetic task prior that defines a
distribution over multi-view learning episodes.
Specifically, each synthetic task $\mathcal{T}$ is formulated as
\begin{equation}
\mathcal{T}
=
\{
\mathcal{S}_{\mathcal{T}},
\mathcal{Q}_{\mathcal{T}}
\}
\sim
p(\mathcal{T}),
\end{equation}
where $\mathcal{S}_{\mathcal{T}}$ and $\mathcal{Q}_{\mathcal{T}}$ denote
support and query sets, respectively. Instead of generating samples from a
fixed distribution, the proposed prior models the variation of multi-view
tasks by jointly considering semantic complexity, view heterogeneity,
cross-view dependency, incomplete observation, and distribution shift.
\subsubsection{Task-level Diversity}

Each task is associated with a specific configuration describing its
classification complexity and multi-view structure. We sample
\begin{equation}
C_{\mathcal T}\sim p_C,
\quad
M_{\mathcal T}\sim p_M,
\quad
D_z\sim p_D,
\end{equation}
where $C_{\mathcal T}$, $M_{\mathcal T}$, and $D_z$ represent the number of classes, available views, and latent semantic dimension, respectively.
The class prior and instance labels are generated as:
\begin{equation}
\boldsymbol{\pi}_{\mathcal T}
\sim
p_{\pi}, \quad y_i
\sim
\operatorname{Categorical}
(\boldsymbol{\pi}_{\mathcal T}).
\end{equation}

The distribution family $p_{\pi}$ covers diverse label structures,
including balanced, imbalanced, and long-tailed scenarios. This enables the
pretrained model to adapt to different class distributions instead of
assuming a fixed classification prior.

For in-context learning, the support set should provide sufficient
information about each class. Therefore, we adopt a constrained sampling
strategy that guarantees class coverage when the support budget permits,
while preserving natural imbalance in the remaining samples.

\subsubsection{Latent Semantic Structure}

We assume that multi-view observations originate from an underlying semantic space shared across views. For each instance, the shared semantic representation is generated from a class-conditional distribution:
\begin{equation}
\mathbf{z}_{i}
\sim
p(\mathbf{z}|y_i,\mathcal T).
\end{equation}
To model complex intra-class variations, we define the conditional
distribution as a mixture of latent modes:
\begin{equation}
k_i
\sim
\operatorname{Categorical}
(\boldsymbol{\rho}_{y_i}),
\end{equation}
\begin{equation}
\mathbf{z}_{i}
=
\boldsymbol{\mu}_{y_i,k_i}
+
\mathbf{L}_{y_i,k_i}
\boldsymbol{\epsilon}_{i},
\quad
\boldsymbol{\epsilon}_{i}
\sim
\mathcal N(0,I).
\end{equation}
Here,
$\boldsymbol{\mu}_{y_i,k_i}$ represents a semantic prototype and
$\mathbf{L}_{y_i,k_i}$ controls the latent covariance structure.
The mixture formulation allows each category to contain multiple semantic
patterns, which better reflects the heterogeneous nature of real-world
multi-view concepts.

To further capture nonlinear dependencies among latent semantic factors, we
introduce a task-specific dependency structure
$\mathbf{A}_{\mathcal T}$:
\begin{equation}
\mathbf{z}_{i}^{*}
=
\mathbf{z}_{i}
+
\alpha_{\mathcal T}
\phi_{\mathcal T}
(
\mathbf{z}_{i}
\mathbf{A}_{\mathcal T}
),
\end{equation}
where $\phi_{\mathcal T}(\cdot)$ denotes a nonlinear transformation and
$\alpha_{\mathcal T}$ controls the dependency strength.
The dependency structure is independently sampled across tasks, producing
diverse latent geometries ranging from nearly independent factors to highly
entangled semantic representations.

\subsubsection{Heterogeneous View Generation}
Although different views describe the same underlying instance, each view provides a distinct observation of the shared semantics. We therefore introduce view-specific private factors:
\begin{equation}
\mathbf{u}_{i}^{(m)}
\sim
p_m(
\mathbf{u}|y_i
).
\end{equation}
The latent representation associated with view $m$ is defined as
\begin{equation}
\mathbf{h}_{i}^{(m)}
=
[
\mathbf{z}_{i};
\mathbf{u}_{i}^{(m)}
].
\end{equation}
The observed view embedding is generated through a
view-specific observation operator:
\begin{equation}
\mathbf{x}_{i}^{(m)}
=
\mathcal O_m
(
\mathbf h_i^{(m)}
)
+
\boldsymbol{\epsilon}_{i}^{(m)},
\end{equation}
where $\mathcal O_m(\cdot)$ represents the view rendering process.
Specifically,
\begin{equation}
\mathcal O_m(\cdot)
=
\mathcal P_m
\circ
g_m
\circ
W_m,
\end{equation}
where $W_m$ controls the projection geometry,
$g_m$ introduces nonlinear transformations, and $\mathcal P_m$ determines
the representation statistics.

By varying these operators, the task prior covers heterogeneous embedding
spaces produced by different pretrained encoders, including differences in
dimensionality, normalization, sparsity, anisotropy, and noise patterns.

\subsubsection{Cross-view Dependency}

Different multi-view tasks exhibit different relationships between views.
Some views contain highly overlapping information, whereas others
provide complementary or noisy evidence.
We therefore control the contribution of shared and private information:
\begin{equation}
\mathbf{x}_{i}^{(m)}
=
\alpha_m
\mathcal O_m^{s}
(\mathbf z_i)
+
(1-\alpha_m)
\mathcal O_m^{p}
(\mathbf u_i^{(m)})
+
\epsilon_i^{(m)},
\end{equation}
where $\alpha_m$ determines the degree of cross-view consistency.
Large $\alpha_m$ produces redundant views, while small $\alpha_m$ generates
more complementary view-specific representations.

\subsubsection{Incomplete Observation and Distribution Shift}
Real-world multi-view systems frequently suffer from missing views and
distribution shifts. We introduce a view availability variable:
\begin{equation}
r_{i,m}
\sim
Bernoulli
(
1-p_{i,m}
),
\end{equation}
where the missing probability is determined by
\begin{equation}
p_{i,m}
=
\sigma
(
f_m(\mathbf z_i)
).
\end{equation}
This formulation allows the missing mechanism to depend on latent semantic
properties, thereby generating realistic missing-not-at-random patterns.
Furthermore, support and query samples may originate from different task
states:
\begin{equation}
\mathbf z_i^{q}
=
\mathbf z_i
+
\Delta_{\mathcal T},
\end{equation}
where $\Delta_{\mathcal T}$ represents task-specific distribution shift.
These variations encourage the model to learn robust multi-view inference
rather than relying on fixed view configurations.

\subsection{Hierarchical Multi-view In-Context Transformer}
\label{subsec:architecture}

The central challenge of multi-view in-context learning is to infer the
relationship between support demonstrations and query instances while
handling heterogeneous views. A direct tokenization strategy that treats
each view representation as an atomic vector ignores the internal
structure of individual views and the hierarchical dependency among
features, views, and instances.
To address this issue, we introduce a hierarchical multi-view Transformer
that performs reasoning at three levels:

1) feature-level representation learning within each view;

2) view-level semantic interaction across views;

3) sample-level in-context reasoning across support and query
instances.

Given an input tensor
$\mathbf{X}
\in
\mathbb{R}^{B\times N\times M\times F},$
where $B$, $N$, $M$, and $F$ denote batch size, number of instances,
number of views, and feature dimension, respectively, the model
constructs a unified task representation through hierarchical
transformations.

\subsubsection{Feature-level Representation Learning}

Different views may originate from heterogeneous pretrained encoders
and therefore exhibit distinct feature organizations. Treating the entire
view embedding as a single token may lose fine-grained feature
dependencies.
Therefore, for each view $m$, we divide the feature representation into
$G$ feature patches:
\begin{equation}
\mathbf{x}_{i}^{(m)}
=
[
\mathbf{x}_{i}^{(m,1)},
...,
\mathbf{x}_{i}^{(m,G)}
].
\end{equation}
Each feature patches is projected into the shared model space:
\begin{equation}
\mathbf{u}_{i}^{(m,g)}
=
\operatorname{MLP}_{f}
(
\mathbf{x}_{i}^{(m,g)}
)
+
\mathbf{e}_{g}^{f},
\end{equation}
where $\mathbf{e}_{g}^{f}$ encodes the relative position of each feature
patches.
The feature-level Transformer captures intra-view dependencies:
\begin{equation}
\{
\widetilde{\mathbf{u}}_{i}^{(m,g)}
\}_{g=1}^{G}
=
\operatorname{Tr}_{f}
(
\{
\mathbf{u}_{i}^{(m,g)}
\}_{g=1}^{G}
).
\end{equation}
The resulting view representation is obtained through attentive
aggregation:
\begin{equation}
\mathbf{h}_{i}^{(m)}
=
\operatorname{LN}
(
\sum_{g=1}^{G}
\alpha_{i,g}^{(m)}
\widetilde{\mathbf{u}}_{i}^{(m,g)}
),
\label{eq:feature_pool}
\end{equation}
where $\alpha_{i,g}^{(m)}$ denotes the learned importance weight of each
feature group.
Compared with simple averaging, the adaptive aggregation enables the model
to emphasize task-relevant feature subspaces.

\subsubsection{View-level Cross-view Reasoning}

After obtaining view-specific representations, the model performs
cross-view reasoning to discover complementary and redundant information
among views.
Each view token is augmented with a view identity embedding:
\begin{equation}
\mathbf{v}_{i}^{(m)}
=
\mathbf{h}_{i}^{(m)}
+
\mathbf{e}_{m}^{\mathrm{mod}} .
\end{equation}
The view Transformer performs self-attention over available views:
\begin{equation}
\{
\widetilde{\mathbf{v}}_{i}^{(m)}
\}_{m=1}^{M}
=
\operatorname{Tr}_{m}
(
\{
\mathbf{v}_{i}^{(m)}
\}_{m=1}^{M};
\mathbf{r}_{i}
),
\end{equation}
where
$\mathbf r_i=[r_{i,1},...,r_{i,M}]$
is the view availability mask. The mask prevents attention from
attending to missing views, enabling the model to dynamically adapt to
different view configurations.
The multi-view representation of instance $i$ is computed as:
\begin{equation}
\mathbf{s}_{i}
=
\operatorname{LN}
\left(
\frac{
\sum_{m=1}^{M}
r_{i,m}
\widetilde{\mathbf{v}}_{i}^{(m)}
}{
\sum_{m=1}^{M}r_{i,m}+\epsilon
}
\right).
\label{eq:modality_fusion}
\end{equation}

This formulation allows the model to aggregate arbitrary subsets of
views without requiring a fixed view combination during
pretraining.

\subsubsection{Sample-level In-context Reasoning}
Following the in-context learning paradigm, prediction is formulated as
reasoning over a task-specific context rather than learning a fixed
classifier.
Given support demonstrations
$\mathcal S=
\{
(\mathbf{s}_i,y_i)
\}_{i=1}^{N_s},$
and query samples
$\mathcal Q=
\{
\mathbf{s}_i
\}_{i=N_s+1}^{N},$
the model jointly processes support and query instances.

For support samples, labels are embedded as additional task context,
whereas query labels are replaced with an unknown token
Each instance token is constructed as:
\begin{equation}
\mathbf{t}_{i}
=
\operatorname{LN}
(
\mathbf{s}_{i}
+
\mathbf{e}_{\widetilde y_i}^{label}
+
\mathbf{e}_{a_i}^{role}
),
\end{equation}
where $a_i$ indicates whether the sample belongs to the support or query
set.
The complete task sequence is processed by the sample-level Transformer:
\begin{equation}
\{
\mathbf{o}_{i}
\}_{i=1}^{N}
=
\operatorname{Tr}_{s}
(
\{
\mathbf{t}_{i}
\}_{i=1}^{N}
),
\label{eq:sample_transformer}
\end{equation}
Through joint attention over support and query samples, each query
representation can dynamically retrieve relevant labeled demonstrations,
which enables task-adaptive prediction without optimizing dataset-specific
parameters.



\subsubsection{Model Pretraining}
  \label{subsec:training}

  For every optimization step, a batch of independent tasks is sampled
  online from the multi-view task prior. The model observes support
  features and labels together with query features, while query labels
  are hidden. The backbone is optimized using query-only
  cross-entropy:
  \begin{equation}
      \mathcal{L}
      =
      -
      \mathbb{E}_{\mathcal{T}\sim p(\mathcal{T})}
      \left[
      \frac{1}{N_{\mathrm{q}}}
      \sum_{i\in\mathcal{D}_{\mathrm{q}}}
      \log
      p_{\theta}
      \left(
      y_i
      \mid
      \mathcal{D}_{\mathrm{s}},
      \mathbf{X}_{\mathrm{q}},
      \mathbf{R}
      \right)
      \right].
      \label{eq:training_loss}
  \end{equation}
  Because view dropout, conditional missingness, observation noise,
  and embedding transformations are resampled online, Eq.~\eqref{eq:training_loss}
  also acts as implicit masked-view and distribution-robust
  pretraining.
For a downstream multi-view dataset, each raw view is encoded using
Eq.~\eqref{eq:modality_encoder}, and each encoder representation is mapped
through Eq.~\eqref{eq:modality_projection}.
We consider two deployment protocols.

\subsubsection{Direct In-Context Inference (SIMPLE-F)}

The view encoders and the inference backbone are frozen. A labeled support
set is provided as context, and query labels are predicted in a single forward
pass without updating the inference backbone.

\subsubsection{Adapter-Only Calibration (SIMPLE-A)}

Only the view adapters and output head are optimized:
\begin{equation}
    \boldsymbol{\theta}_{\mathrm{adapt}}
    =
    \left\{
        \boldsymbol{\theta}_{P_1},
        \ldots,
        \boldsymbol{\theta}_{P_M},
        \mathbf{W}_{o},
        \mathbf{b}_{o}
    \right\}.
    \label{eq:adapter_parameters}
\end{equation}

The parameters of the factorized inference transformer remain frozen.

\section{Experiments}
\label{sec:experiments}

We conduct extensive experiments to evaluate the effectiveness and generalization capability of SIMPLE.
Specifically, we aim to answer the following research questions:

\begin{itemize}
    \item \textbf{RQ1:} How does SIMPLE compare with existing multi-view learning methods on standard benchmarks?
    \item \textbf{RQ2:} Can SIMPLE generalize to unseen tasks with limited supervision and distribution shifts?
    \item \textbf{RQ3:} How robust is SIMPLE when some views are unavailable during inference?
    \item \textbf{RQ4:} How does each component contribute to the final performance?
\end{itemize}

\subsection{Experimental Setup}
\label{subsec:experimental_setup}

\subsubsection{Datasets}

We evaluate SIMPLE on two benchmark suites, including multi-view
classification and multi-omics classification datasets.
The multi-view suite contains seven datasets:
HW, OutScene, ESPGame, Flickr, NUSWIDE, YouTube, and Flowers.
The multi-omics suite includes six cancer subtype datasets:
GS-BRCA, GS-COAD, GS-GBM, GS-LGG, GS-OV, and Pan-Cancer \cite{yang2025mlomics}.
Details are provided in the Appendix.

\subsubsection{Compared Methods}

We compare SIMPLE with representative baselines from different
categories, including conventional predictors, multi-view representation
learning methods, and
multi-omics integration methods.
The compared methods include MLP, SVM,
XGBoost~\cite{chen2016xgboost}, DCCA~\cite{dcca},
DMF~\cite{wang2015deep}, Co-GCN~\cite{li2020co},
RCML~\cite{xu2024reliable}, TUNED~\cite{huang2024trusted},
ECMGD~\cite{ECMGD}, ViHMGD~\cite{lu2025views},
DeepMO~\cite{DeepMo}, MOGONET~\cite{wang2021mogonet},
MoGCN~\cite{Li2022MOGCN}, and GTMancer~\cite{gtmancer}.
Detailed descriptions of all baselines are provided in the Appendix.



\begin{table*}[!t]
\centering
\renewcommand{\arraystretch}{1.15}

\resizebox{\textwidth}{!}{%
\begin{tabular}{lcccccccccccccc}
\toprule

\multirow{2}{*}{Method}
& \multicolumn{2}{c}{HW}
& \multicolumn{2}{c}{OutScene}
& \multicolumn{2}{c}{Espgame}
& \multicolumn{2}{c}{Flickr}
& \multicolumn{2}{c}{NUSWIDE}
& \multicolumn{2}{c}{Youtube}
& \multicolumn{2}{c}{Flowers} \\

\cmidrule(lr){2-3}
\cmidrule(lr){4-5}
\cmidrule(lr){6-7}
\cmidrule(lr){8-9}
\cmidrule(lr){10-11}
\cmidrule(lr){12-13}
\cmidrule(lr){14-15}

& ACC & F1
& ACC & F1
& ACC & F1
& ACC & F1
& ACC & F1
& ACC & F1
& ACC & F1 \\

\midrule

\rowcolor{gray!15}
\multicolumn{15}{c}{\textit{Conventional Methods}} \\

MLP
& 93.1 & 93.1
& 79.7 & 80.0
& 83.2 & 84.0
& 68.0 & 67.9
& 44.1 & 44.1
& 66.8 & 66.4
& 60.4 & 60.1 \\

DCCA
& 94.8 & 94.8
& \underline{80.5} & \underline{80.7}
& 83.1 & 82.8
& 69.3 & 69.0
& 45.1 & 45.2
& 70.9 & 70.6
& 59.5 & 58.6 \\

DMF
& 94.2 & 94.2
& 80.1 & 80.4
& 85.8 & \underline{85.6}
& \underline{70.0} & 69.9
& 45.2 & \underline{45.2}
& {70.9} & {70.6}
& \underline{67.7} & \underline{67.8} \\

\midrule

\rowcolor{gray!15}
\multicolumn{15}{c}{\textit{Multi-view Fusion Methods}} \\

CoGCN
& 91.6 & 86.9
& 71.0 & 71.3
& 75.9 & 75.5
& 61.2 & 61.1
& 40.6 & 37.0
& 29.3 & 21.5
& 28.4 & 25.6 \\

RCML
& 92.4 & 92.4
& 78.9 & 79.1
& \textbf{86.0} & \textbf{85.8}
& 70.9 & 70.4
& 42.5 & 41.3
& \textbf{71.2} & \textbf{71.9}
& 10.6 & 3.2 \\

TUNED
& 90.2 & 90.1
& 77.3 & 77.5
& 84.5 & 84.2
& 68.9 & 68.4
& 43.0 & 40.6
& 66.0 & 65.4
& 5.9 & 0.7 \\

\midrule

\rowcolor{gray!15}
\multicolumn{15}{c}{\textit{Transformer Models}} \\

ECMGD
& \underline{95.6} & \underline{95.6}
& 79.3 & 79.3
& 84.3 & 84.0
& 70.5 & \underline{70.4}
& \textbf{47.5} & 46.5
& 59.4 & 59.0
& 61.0 & 60.8 \\

ViHMGD
& 91.7 & 91.8
& 71.1 & 71.5
& 85.0 & 84.7
& 66.0 & 65.8
& 46.4 & 46.5
& 57.4 & 56.9
& 11.7 & 5.4 \\

\midrule

SIMPLE-F
& 93.4 & 93.4
& 75.0 & 74.8
& 69.8 & 69.6
& 61.6 & 61.7
& 43.8 & 42.3
& 55.7 & 54.5
& 61.9 & 61.0 \\

SIMPLE-A
& \textbf{96.2} & \textbf{96.2}
& \textbf{81.9} & \textbf{81.3}
& \underline{85.9} & 85.3
& \textbf{71.4} & \textbf{71.3}
& \underline{46.1} & \textbf{46.4}
& \underline{71.0} & \underline{71.1}
& \textbf{69.4} & \textbf{69.0} \\

\bottomrule
\end{tabular}
}
\caption{Performance comparison on multi-view datasets. The best available
results are highlighted in bold and the second-best available results are
underlined. Each entry reports mean ACC and macro-F1 scores (\%) over five runs.}
\label{tab:view}
\end{table*}

\subsection{Overall Comparison}
\label{subsec:overall}
\subsubsection{Classification (RQ1)}
We first evaluate the overall performance of \method on diverse multi-view benchmarks, including multi-view classification and multi-omics prediction tasks. More results are shown in Appendix.
Tables~\ref{tab:view} and~\ref{tab:omics} summarize the comparison results on multi-view and multi-omics benchmarks, respectively.
Several observations can be drawn from the results. 
First, SIMPLE-F achieves competitive performance against fully trained multi-view learning approaches, despite requiring no task-specific optimization.
This demonstrates that the proposed synthetic multi-view task pretraining enables the model to capture transferable inference patterns rather than relying on dataset-specific correlations.
Second, after lightweight adaptation, SIMPLE-A consistently improves upon SIMPLE-F and achieves superior performance on most benchmarks.
On the multi-view datasets, SIMPLE-A obtains the best results on HW, OutScene, Flickr, and Flowers, achieving accuracies of 96.2\%, 81.9\%, 71.4\%, and 69.4\%, respectively.
Similarly, on multi-omics benchmarks, SIMPLE-A achieves leading performance on multiple cancer subtype classification tasks, demonstrating its effectiveness in modeling highly heterogeneous biological views.

The consistent improvement from SIMPLE-F to SIMPLE-A reveals an important property of the proposed framework: the pretrained model already learns general multi-view reasoning ability, while lightweight adaptation further aligns the learned inference process with task-specific characteristics.

\begin{table*}[t]
\centering

\renewcommand{\arraystretch}{1.15}

\begin{tabular}{lccccccccccccc}
\toprule

\multirow{2}{*}{Method}
& \multicolumn{2}{c}{GS-BRCA}
& \multicolumn{2}{c}{GS-COAD}
& \multicolumn{2}{c}{GS-GBM}
& \multicolumn{2}{c}{GS-LGG}
& \multicolumn{2}{c}{GS-OV}
& \multicolumn{2}{c}{Pan-Cancer} \\

\cmidrule(lr){2-3}
\cmidrule(lr){4-5}
\cmidrule(lr){6-7}
\cmidrule(lr){8-9}
\cmidrule(lr){10-11}
\cmidrule(lr){12-13}

& ACC & F1
& ACC & F1
& ACC & F1
& ACC & F1
& ACC & F1
& ACC & F1 \\

\midrule

\rowcolor{gray!10}
\multicolumn{13}{c}{\textit{Single-view and Statistical Methods}} \\

SVM
& \underline{74.9} & 53.0
& 80.4 & 47.8
& 60.1 & 57.0
& 94.2 & 93.4
& 66.0 & 65.3
& 94.6 & {88.3} \\

XGBoost
& 74.7 & 49.9
& 79.2 & 42.2
& 58.7 & 53.6
& 91.6 & 89.9
& 67.0 & 65.5
& 93.4 & 85.6 \\

MLP
& 71.4 & 58.6
& 79.2 & 58.4
& 61.3 & {62.1}
& 93.0 & 92.3
& 64.9 & 64.5
& \underline{95.0} & \underline{90.3} \\

\midrule

\rowcolor{gray!10}
\multicolumn{13}{c}{\textit{Multi-omics Representation Learning Methods}} \\

DeepMO
& \underline{76.4} & \underline{66.1}
& 80.0 & \textbf{60.1}
& \underline{62.9} & \textbf{63.6}
& 92.6 & 91.7
& 66.2 & 65.6
& 92.0 & 83.9 \\

MOGONET
& 69.7 & 44.1
& 69.7 & 35.1
& 40.0 & 35.0
& 82.9 & 79.5
& 46.3 & 43.3
& 38.3 & 16.7 \\

MoGCN
& 70.3 & 53.2
& 79.6 & 46.6
& 56.5 & 57.3
& 92.1 & 90.8
& 67.0 & 66.2
& 89.9 & 81.1 \\

GTMancer
& 70.6 & 42.7
& \underline{81.1} & 44.1
& 58.0 & 50.3
& 92.8 & 90.8
& 54.9 & 51.4
& 92.8 & 86.1 \\

ViHMGD
& 65.1 & 53.3
& 44.9 & 36.5
& 57.3 & 54.7
& 91.6 & 90.9
& \textbf{68.4} & \underline{67.2}
& 88.2 & 74.2 \\

\midrule

SIMPLE-F
& 71.7 & 64.5
& 70.5 & 47.8
& 54.6 & 55.5
& \underline{95.1} & \underline{94.0}
& 62.9 & 62.5
& 86.6 & 82.3 \\

SIMPLE-A
& \textbf{77.7} & \textbf{68.5}
& \textbf{82.1} & \underline{59.5}
& \textbf{63.4} & \underline{63.6}
& \textbf{96.9} & \textbf{95.7}
& \underline{67.9} & \textbf{67.9}
& \textbf{95.7} & \textbf{91.7} \\

\bottomrule
\end{tabular}
\caption{Performance comparison on multi-omics datasets. The best available
results are highlighted in bold and the second-best available results are
underlined. Each entry reports mean ACC and macro-F1 scores (\%) over five runs.}
\label{tab:omics}
\end{table*}
\begin{figure*}[!t]
\centering
\begin{minipage}[t]{0.49\textwidth}
\centering
\includegraphics[width=\linewidth]{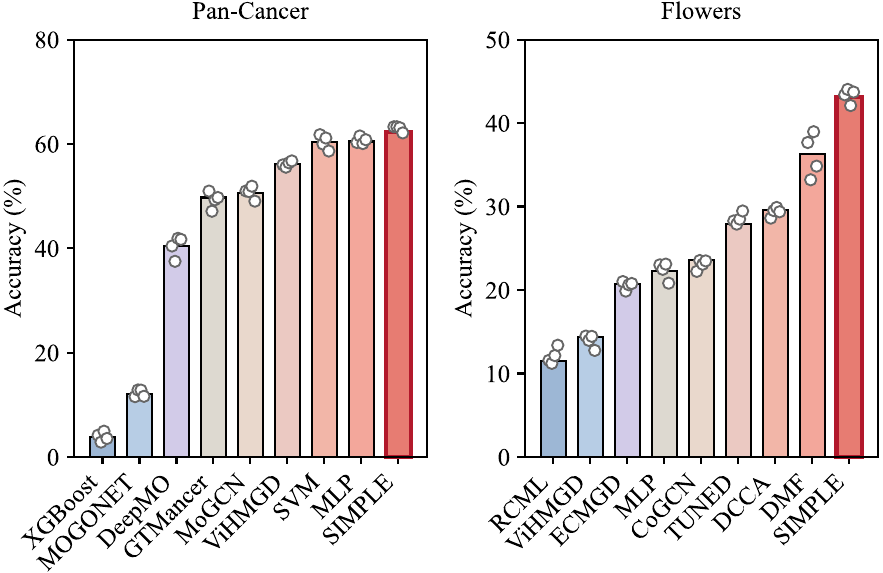}
\caption{Performance comparison under the one-shot learning setting on different datasets.}
\label{oneshot}
\end{minipage}
\hfill
\begin{minipage}[t]{0.49\textwidth}
\centering
\includegraphics[width=\linewidth]{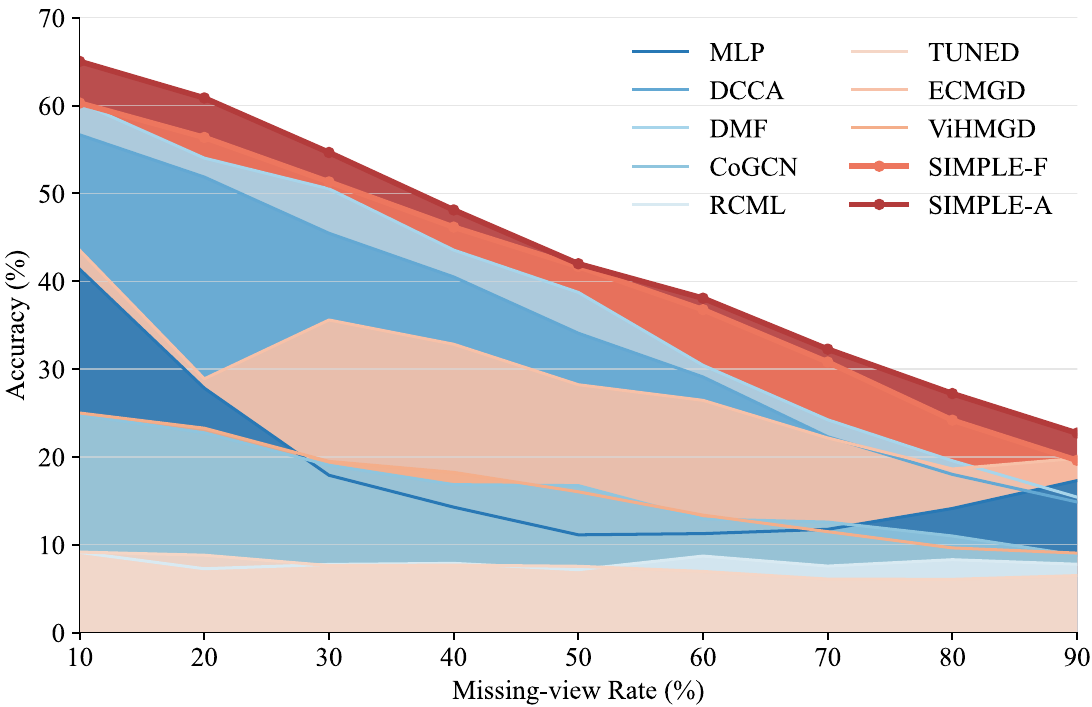}
\caption{Robustness analysis under different missing-view ratios on the Flowers dataset.}
\label{miss}
\end{minipage}
\end{figure*}
\subsubsection{One-shot Learning (RQ2)}
To evaluate the capability of \method under extremely limited supervision, we conduct one-shot learning experiments, where only one labeled sample per class is available during training.
As shown in Fig.~\ref{oneshot}, all baseline methods experience significant performance degradation due to insufficient labeled examples.
In contrast, \method consistently achieves superior performance on both Pan-Cancer and Flowers datasets.
These results indicate that \method can effectively exploit task-level prior knowledge and contextual information, enabling robust inference beyond conventional supervised learning settings.

\subsubsection{Robustness to Missing Views (RQ3)}

Real-world multi-view systems often suffer from incomplete observations.
Therefore, we evaluate the robustness of \method by randomly removing views with different missing ratios on the Flowers dataset.
As shown in Fig.~\ref{miss}, all methods exhibit performance degradation as the missing-view ratio increases.
However, \method maintains a slower degradation rate compared with existing approaches.
Especially under severe missing-view conditions, \method preserves higher classification accuracy, demonstrating its ability to infer reliable predictions from incomplete multi-view information.
This advantage can be attributed to the proposed adaptive view reasoning mechanism, which dynamically exploits available views instead of relying on fixed view combinations.

\begin{figure*}[!t]
\centering
\begin{minipage}[t]{0.49\textwidth}
\centering
\includegraphics[width=\linewidth]{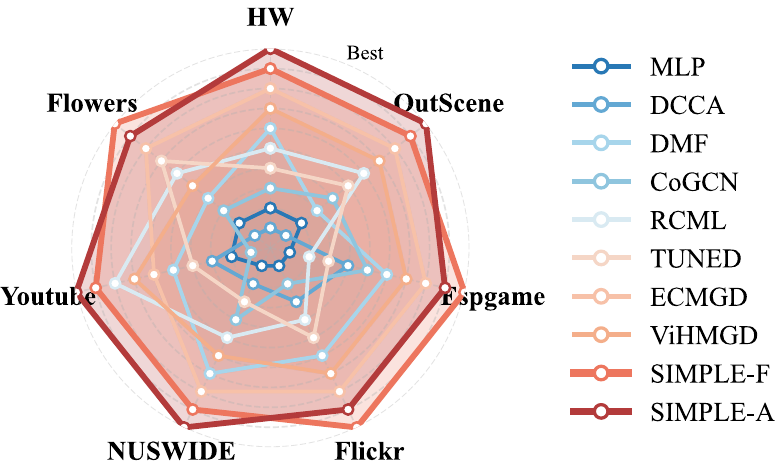}
\caption{Cross-dataset ranking comparison after training on Pan-Cancer.}
\label{Cross}
\end{minipage}
\hfill
\begin{minipage}[t]{0.49\textwidth}
\centering
\includegraphics[width=\linewidth]{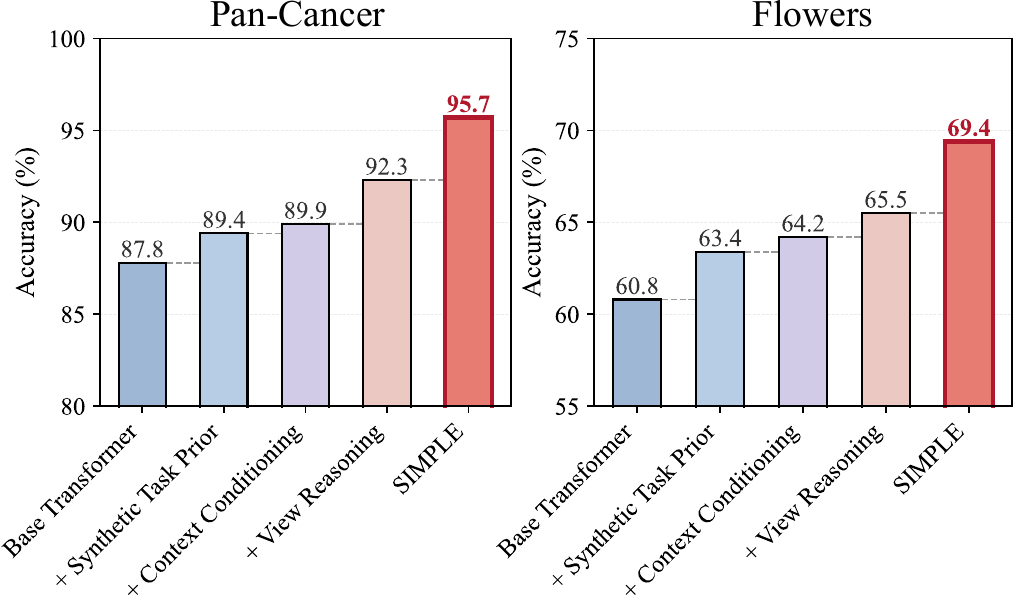}
\caption{Ablation study of different components in \mbox{SIMPLE}.}
\label{Ablation}
\end{minipage}
\end{figure*}

\subsubsection{Cross-dataset Transfer (RQ2)}

To complement the task-level transfer evaluated by \method-F, we conduct a cross-domain backbone-transfer experiment.
All methods are trained on the Pan-Cancer dataset. For each unseen target dataset, the source-trained backbone is frozen and only a target-side linear classifier is fitted from target support data; full protocol details are provided in the \emph{Cross-domain Transfer Evaluation} appendix section.
This protocol evaluates transfer across domains and label spaces and should not be interpreted as parameter-free contextual inference.
Instead of reporting absolute accuracy, Fig.~\ref{Cross} summarizes the relative ranking of different methods across target datasets, providing a more comprehensive view of their transferability under heterogeneous domain shifts.
As shown in Fig.~\ref{Cross}, SIMPLE consistently maintains favorable rankings across diverse target datasets. These results provide complementary evidence that source-trained representations remain useful under heterogeneous domain shifts.

\subsubsection{Ablation Study (RQ4)}
To investigate the contribution of each component in \method, we conduct
progressive ablation experiments by gradually adding each proposed module.
As shown in Fig.~\ref{Ablation}, all components consistently improve the
performance over the base Transformer on  Pan-Cancer and Flowers
datasets. Specifically, introducing the synthetic task prior equips the
model with inference regularities learned across diverse multi-view tasks,
while context conditioning enables those regularities to be instantiated
for a new task. The additional gains from view reasoning
demonstrate the effectiveness of explicitly modeling interactions among
different views. 
These results verify that each component contributes independently and
their combination provides complementary benefits for effective
multi-view inference.

\section{Conclusion}
In this work, we reconsidered multi-view learning from the perspective of reusable inference. While pretrained encoders make view-specific representations increasingly transferable, existing methods still relearn how to interpret and combine these views for every downstream dataset. To address this limitation, we proposed \method that amortizes task-specific fusion and prediction into pretraining. A controllable synthetic task prior exposes the model to diverse learning problems with varying class structures, representation geometries, cross-view relationships, view reliability, missingness patterns, and distribution shifts. Based on these tasks, the hierarchical inference architecture performs reasoning within individual views, across available views, and between support and query samples.
Experiments on multi-view and multi-omics benchmarks validate both the transferability and adaptability of the learned inference procedure. These findings suggest that multi-view reasoning can be amortized across tasks, rather than discarded and relearned from scratch for every downstream dataset.

\bibliographystyle{unsrt}
\bibliography{aaai2027}

\newpage
\appendix

\section{Detailed Experimental Settings}
\label{app:experimental_details}

This appendix provides additional details about the datasets, compared
methods, evaluation protocols, implementation settings, and transfer
experiments. Unless otherwise specified, all downstream datasets are
strictly excluded from synthetic pretraining.

\subsection{Datasets}
\label{app:datasets}

We evaluate \method on 13 datasets, including seven general multi-view
classification datasets and six multi-omics cancer classification
datasets. For every dataset, multiple views describing the same sample
share a common classification label.

\subsubsection{General Multi-view Datasets}

\begin{table*}[t]
\centering
\caption{
Statistics of the general multi-view classification datasets.
Flower17 is denoted as Flowers in the main paper.
}
\label{tab:app_multiview_datasets}
\small
\setlength{\tabcolsep}{4.5pt}
\renewcommand{\arraystretch}{1.10}
\begin{tabular}{lrrrrp{7.0cm}}
\toprule
Dataset & Samples & Classes & Views & Domain & View dimensions \\
\midrule
HW
& 2,000 & 10 & 6 & Handwritten digits
& 153, 596, 301, 481, 157, and 27 \\

OutScene
& 2,688 & 8 & 4 & Outdoor scenes
& 512, 432, 256, and 48 \\

ESP-Game
& 11,032 & 7 & 2 & Image--text
& 100 and 100 \\

Flickr
& 12,154 & 6 & 2 & Image--text
& 100 and 100 \\

NUSWIDE
& 1,600 & 8 & 6 & Web images
& 64, 144, 73, 128, 225, and 500 \\

YouTube
& 2,000 & 10 & 6 & Video and audio
& 2,000, 1,024, 64, 512, 64, and 647 \\

Flower17
& 1,360 & 17 & 7 & Flower images
& 1,360 dimensions for each view \\
\bottomrule
\end{tabular}
\end{table*}

\paragraph{HW.}
HW contains 2,000 handwritten digit samples from 10 classes.
Each sample is represented by six visual feature views describing
complementary shape, texture, and statistical properties.

\paragraph{OutScene.}
OutScene contains 2,688 outdoor-scene images from eight semantic classes.
Each image is represented by four visual feature views encoding global
and local image characteristics.

\paragraph{ESP-Game and Flickr.}
ESP-Game and Flickr are image--text datasets. Each sample contains one
visual representation and one textual representation, providing
heterogeneous semantic evidence for classification.

\paragraph{NUSWIDE.}
NUSWIDE contains heterogeneous visual and semantic descriptors of web
images. Its six views differ substantially in dimensionality and feature
statistics.

\paragraph{YouTube.}
YouTube contains video samples represented by six heterogeneous visual
and audio feature views.

\paragraph{Flower17.}
Flower17 contains 1,360 flower images from 17 categories. Each image is
represented by seven visual descriptors. We denote this dataset as
Flowers in the main paper.

\subsubsection{Multi-omics Datasets}

The multi-omics datasets are obtained from the MLOmics benchmark
\cite{yang2025mlomics}. Each patient is represented by four molecular
views: gene expression, DNA methylation, microRNA expression, and
copy-number variation.

\begin{table*}[t]
\centering
\caption{
Statistics of the multi-omics datasets. The four view dimensions
correspond to gene expression, DNA methylation, microRNA expression,
and copy-number variation, respectively.
}
\label{tab:app_omics_datasets}
\small
\setlength{\tabcolsep}{5pt}
\renewcommand{\arraystretch}{1.10}
\begin{tabular}{lrrrp{8.2cm}}
\toprule
Dataset & Patients & Classes & Views & View dimensions \\
\midrule
GS-BRCA
& 671 & 5 & 4
& 11,203; 11,189; 11,343; 310 \\

GS-COAD
& 260 & 4 & 4
& 11,203; 11,189; 11,343; 286 \\

GS-GBM
& 244 & 5 & 4
& 11,205; 11,192; 11,346; 325 \\

GS-LGG
& 247 & 3 & 4
& 11,205; 11,191; 11,345; 328 \\

GS-OV
& 284 & 4 & 4
& 11,205; 11,191; 11,344; 321 \\

Pan-Cancer
& 8,314 & 32 & 4
& 3,105; 3,139; 3,217; 383 \\
\bottomrule
\end{tabular}
\end{table*}

The five GS datasets correspond to cancer-subtype classification within
individual cancer cohorts. Pan-Cancer covers 32 cancer types and exhibits
greater class diversity and distributional heterogeneity.

\subsection{Compared Methods}
\label{app:compared_methods}

We compare \method with conventional classifiers, deep multi-view
representation learning methods, uncertainty-aware fusion methods,
Transformer-based models, and specialized multi-omics integration
approaches.

\subsubsection{Conventional and Representation Learning Methods}

\paragraph{MLP.}
MLP concatenates all available views and feeds the resulting
representation into a two-layer classifier. It serves as a direct
feature-level fusion baseline.

\paragraph{SVM.}
SVM is trained on concatenated and normalized multi-view features using
an RBF kernel. It is evaluated on the multi-omics datasets.

\paragraph{XGBoost.}
XGBoost \cite{chen2016xgboost} is trained on concatenated multi-view
features and provides a non-neural statistical baseline.

\paragraph{DCCA.}
DCCA \cite{dcca} employs view-specific neural encoders and maximizes the
correlation between their latent representations before classification.

\paragraph{DMF (our implementation).}
Our DMF implementation follows the general formulation of deep
multi-view representation learning \cite{wang2015deep}. Each view has an
independent encoder and decoder, while the shared representation is
obtained by averaging the view-specific latent factors. The objective is
defined as
\begin{equation}
\mathcal{L}_{\mathrm{DMF}}
=
\mathcal{L}_{\mathrm{CE}}
+
0.5\mathcal{L}_{\mathrm{rec}}
+
0.1\mathcal{L}_{\mathrm{con}},
\end{equation}
where $\mathcal{L}_{\mathrm{CE}}$, $\mathcal{L}_{\mathrm{rec}}$, and
$\mathcal{L}_{\mathrm{con}}$ denote classification, reconstruction, and
cross-view consistency losses, respectively. Since an official
implementation matching our experimental interface was unavailable, this
baseline is explicitly denoted as \emph{DMF (our implementation)}.

\subsubsection{Multi-view Fusion Methods}

\paragraph{Co-GCN.}
Co-GCN \cite{li2020co} constructs a graph for each view and applies graph
convolution to capture view-specific neighborhood structures.

\paragraph{RCML.}
RCML \cite{xu2024reliable} performs reliability-aware evidential fusion
and reduces the influence of conflicting or uncertain views.

\paragraph{TUNED.}
TUNED \cite{huang2024trusted} combines neighborhood graph construction
with uncertainty-aware evidential learning for incomplete or inconsistent
multi-view observations.

\subsubsection{Transformer-based Methods}

\paragraph{ECMGD.}
ECMGD \cite{ECMGD} employs Transformer-based cross-view interaction to
model view-specific information and shared dependencies.

\paragraph{ViHMGD.}
ViHMGD \cite{lu2025views} combines intra-view representation learning and
inter-view interaction to capture high-order relationships among views.

\subsubsection{Multi-omics Integration Methods}

\paragraph{DeepMO.}
DeepMO \cite{DeepMo} learns nonlinear representations from multiple omics
measurements and integrates them for cancer classification.

\paragraph{MOGONET.}
MOGONET \cite{wang2021mogonet} constructs one patient-similarity graph for
each omics view and combines graph convolution with view-correlation
discovery.

\paragraph{MoGCN.}
MoGCN \cite{Li2022MOGCN} integrates omics-specific representation
learning with graph convolution. We use the provided
\texttt{MOGCN\_main} training entry point.

\paragraph{GTMancer.}
GTMancer \cite{gtmancer} combines graph and Transformer modules to model
sample-level relationships and interactions among omics views.

\subsection{Evaluation Protocol}
\label{app:protocol}

\paragraph{Support--query split.}
For standard classification experiments, we construct a
class-proportional support set containing 10\% of all samples and use the
remaining 90\% as a disjoint query set. At least one labeled support
sample is retained for each class. For a given seed, all methods use the
same support--query split.

\paragraph{Random seeds.}
All experiments are repeated using seeds 42, 43, 44, 45, and 46.
We report the arithmetic mean over the five runs. When uncertainty is
reported, we use the sample standard deviation
\begin{equation}
s
=
\sqrt{
\frac{1}{n-1}
\sum_{i=1}^{n}
(x_i-\bar{x})^2
}.
\end{equation}

\paragraph{Evaluation metrics.}
We report classification accuracy and macro-averaged F1. Macro-F1 gives
equal weight to each class and is therefore suitable for datasets with
class imbalance.

\paragraph{Query-label isolation.}
Query labels are never used for training, adapter calibration,
normalization, checkpoint selection, or hyperparameter selection.
All normalization statistics and trainable downstream parameters are
estimated exclusively from the support set.

\paragraph{Baseline training.}
MLP, DCCA, DMF, Co-GCN, RCML, TUNED, ECMGD, and ViHMGD are trained for 300 epochs.
Whenever an official implementation is available, its original model and
loss definitions are retained.

\section{Implementation Details of SIMPLE}
\label{app:simple_settings}

\subsection{Input Processing}

For \method, normalization statistics are estimated from the support set
only and subsequently applied to both support and query samples. To handle
heterogeneous feature dimensions, each view is mapped into a
256-dimensional space using a fixed Gaussian random projection:
\begin{equation}
\mathbf{R}_{ij}
\sim
\mathcal{N}
\left(
0,
\frac{1}{d_{\mathrm{in}}}
\right).
\end{equation}

The projection matrix is sampled once for each run and remains fixed for
all support and query samples in that run. Missing views are zero-filled
for tensor construction but are excluded from attention and aggregation
using an explicit binary availability mask.

\subsection{Synthetic Multi-view Task Prior}

\method is pretrained exclusively on procedurally generated multi-view
classification tasks. No downstream sample, label, or dataset statistic
is used during synthetic pretraining.

The main task-prior settings are summarized as follows:

\begin{itemize}
    \item The number of views is sampled from 1 to 16.
    \item The number of classes is sampled from 2 to 64.
    \item The latent semantic dimension is sampled from 8 to 128.
    \item Each task contains up to 4,096 support samples and 1,024 query
    samples.
    \item Each class may contain up to 10 latent mixture components.
    \item The prior generates shared, view-private, redundant,
    complementary, weak, uninformative, and conflicting information.
    \item Observation functions include linear, hyperbolic-tangent,
    sinusoidal, SiLU, signed-square-root, and quadratic transformations.
    \item Representations may be anisotropic, sparse, low-rank,
    correlated, quantized, heavy-tailed, count-like, or
    $\ell_2$-normalized.
    \item The prior includes class imbalance, noisy labels, outliers,
    spurious correlations, support--query shifts, conditional
    missingness, block missingness, and whole-view dropout.
    \item At least one view is retained for every generated sample.
\end{itemize}

The synthetic prior generates distinct learning problems rather than
additional samples from a fixed problem. Its purpose is to expose the
inference model to broad variations in task structure, representation
geometry, cross-view dependency, and observation conditions.

\subsection{Model Architecture}

The feature-hybrid \method backbone contains 125,005,122 trainable
parameters. Its main architectural settings are:

\begin{itemize}
    \item Each 256-dimensional input view is divided into four
    64-dimensional feature groups.
    \item A shared MLP maps every feature group to a 768-dimensional token.
    \item One 12-head Transformer layer performs feature-level interaction
    within each view.
    \item Four 12-head Transformer layers model interactions among the
    available views.
    \item A mask-aware normalized sum aggregates the available views.
    \item Support samples receive task-local label embeddings, while
    support and query samples receive distinct role embeddings.
    \item A 14-layer, 12-head Transformer performs contextual reasoning
    across support and query samples.
    \item All Transformer blocks use pre-normalization, residual
    connections, GELU feed-forward networks, and dropout of 0.1.
    \item The prediction decoder combines a learned classifier with
    cosine-similarity logits computed from contextualized support
    prototypes.
\end{itemize}

\subsection{Pretraining Settings}

\method is optimized using AdamW with learning rate $10^{-4}$, weight
decay $10^{-4}$, and gradient-norm clipping at 1.0. Each optimization
step contains four independently sampled synthetic tasks. Pretraining uses
bfloat16 automatic mixed precision and TF32 matrix multiplication. The
training seed is 42.

\subsection{Frozen Inference}

\method-F directly applies the pretrained inference model to a downstream
task. The labeled support samples and unlabeled query samples are jointly
processed in the same context, and no model parameter is updated.
This protocol evaluates whether the learned multi-view inference
procedure transfers directly to an unseen dataset.

\subsection{Adapter-only Calibration}

\method-A freezes the complete inference backbone and optimizes only the
following lightweight parameters:

\begin{itemize}
    \item input LayerNorm parameters;
    \item view-specific scale and bias parameters;
    \item a low-rank residual adapter;
    \item a scalar logit temperature; and
    \item per-class logit biases.
\end{itemize}

For the general multi-view datasets, we use adapter rank 32, 30 AdamW
updates, learning rate $3\times10^{-3}$, weight decay $10^{-4}$, and
gradient clipping at 1.0. For the multi-omics datasets, we use adapter
rank 2, 80 updates, and learning rate $10^{-3}$.

Adapter calibration uses the support set only. For each eligible class,
70\% of the support samples form the in-context subset, while the
remaining 30\% are treated as pseudo-query samples. Real query labels are
never accessed.

\subsection{One-shot Evaluation}

For one-shot evaluation, exactly one labeled support sample is selected
from every class, and all remaining samples form the query set. All
methods use the same support samples for a given seed.

\subsection{Missing-view Evaluation}

For a nominal missing rate between 10\% and 90\%, each sample--view pair
is independently selected for removal. If all views of a sample are
removed, one view is randomly restored so that every sample retains at
least one observation.

The masks are nested across missing rates. Therefore, a view removed at a
lower missing rate remains unavailable at all higher missing rates.
For each dataset, missing rate, and seed, all methods use exactly the same
mask realization.

\section{Cross-domain Transfer Evaluation}
\label{app:cross_domain}

We further evaluate whether a representation backbone learned from the
multi-omics domain can be reused on general multi-view datasets.
Pan-Cancer is used as the source dataset, while HW, OutScene, ESP-Game,
Flickr, NUSWIDE, YouTube, and Flowers are treated as unseen target
datasets.

This experiment should be interpreted as frozen-backbone transfer with
target-side linear probing rather than zero-shot classification. Since
source and target datasets have different label spaces, a new linear
classifier is trained on the target support set, while the representation
backbone remains frozen.

\subsection{Parameter-free Input Alignment}

The source and target datasets have different feature dimensions and
numbers of views. We therefore apply a parameter-free alignment procedure
that does not use target labels.

For each sample and raw view, feature values are standardized using their
sample-wise mean and standard deviation, clipped to $[-5,5]$, sorted, and
linearly interpolated at 128 uniformly spaced quantile positions. This
operation converts each raw view into a 128-dimensional distributional
descriptor.

The number of views is then converted to four:

\begin{itemize}
    \item For two-view datasets, we use the two original descriptors,
    their element-wise mean, and their absolute difference.
    \item For three-view datasets, we use the three original descriptors
    and their element-wise mean.
    \item Four-view datasets are kept unchanged.
    \item Datasets with more than four views are divided into four groups,
    and descriptors within each group are averaged.
\end{itemize}

The same procedure is applied to Pan-Cancer and every target dataset. It
contains no trainable parameter and uses no target labels.

\subsection{Source Training and Target Linear Probe}

For each compared baseline, the transfer experiment consists of four
stages:

\begin{enumerate}
    \item Train the method on all labeled Pan-Cancer samples for 50 epochs.
    \item Freeze the complete representation backbone after source
    training.
    \item Initialize a new linear classifier for each target dataset and
    train it using a class-proportional 10\% target support set for
    300 epochs.
    \item Evaluate the frozen backbone and trained linear classifier on
    the disjoint 90\% target query set.
\end{enumerate}

The target linear classifier is optimized using AdamW with learning rate
$10^{-2}$, weight decay $10^{-4}$, and class-weighted cross-entropy.
Before linear-probe training, target representations are standardized
using statistics estimated from the target support set only.

\subsection{Rank-based Visualization}

The radar plot reports relative ranks rather than absolute classification
accuracy. This choice avoids directly comparing raw accuracy values across
target datasets with different numbers of classes and difficulty levels.

For each target dataset $d$, methods are ranked according to their mean
classification accuracy. Rank one denotes the best-performing method.
Tied methods receive their average rank. For visualization, we convert the
rank into a normalized score:
\begin{equation}
r_{m,d}
=
\frac{K-q_{m,d}}{K-1},
\end{equation}
where $q_{m,d}$ denotes the rank of method $m$ on dataset $d$, and $K$
is the number of compared methods. A larger value therefore indicates a
better relative rank.

The radar plot should be interpreted as a comparison of ranking
consistency across heterogeneous target datasets, rather than a
visualization of absolute performance.

\section{Additional Efficiency Analysis}
\label{app:efficiency}

\subsection{Measurement Protocol}

We compare predictive accuracy, downstream execution time, and peak
allocated GPU memory. All methods are measured using the same hardware,
software environment, downstream split, and random seed.

Let $t$ denote the measured downstream execution time in seconds. The
horizontal axis reports $\log_{2}(1+t)$ to accommodate large runtime
differences across methods. Bubble size represents peak allocated GPU
memory measured using
\texttt{torch.cuda.max\_memory\_allocated}.

For fully trained baselines, the execution time includes task-specific
training. For \method-A, it includes adapter calibration. For
\method-F, it corresponds to direct contextual inference without
downstream optimization. The figure therefore compares downstream
adaptation and inference cost rather than only single-batch inference
latency.

\begin{figure}[t]
\centering
\includegraphics[width=0.5\linewidth]{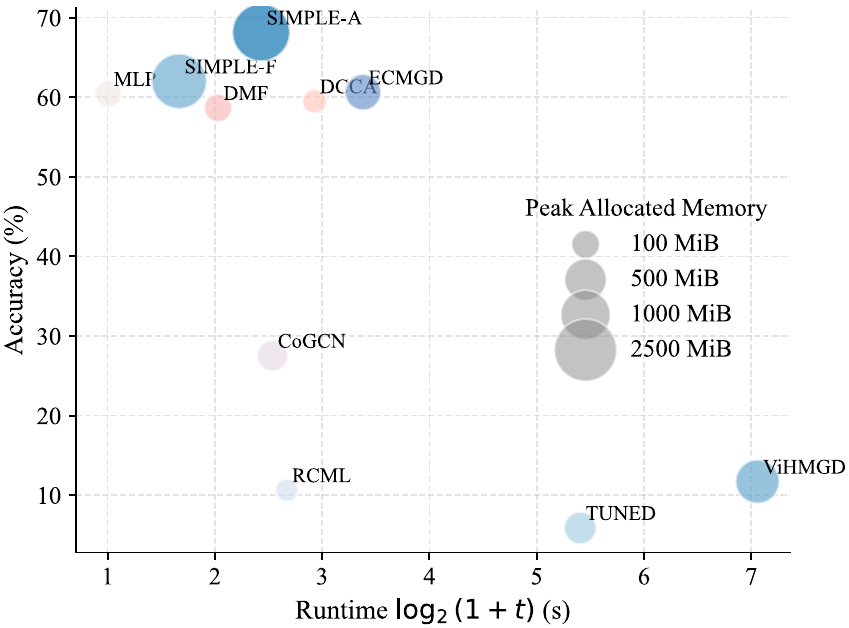}
\caption{
Accuracy--efficiency comparison of different methods.
The horizontal axis reports $\log_{2}(1+t)$, where $t$ is the downstream
execution time in seconds. The vertical axis reports classification
accuracy, while bubble size represents peak allocated GPU memory.
Higher, further-left, and smaller points are preferable.
}
\label{fig:app_efficiency}
\end{figure}

\subsection{Efficiency Results}

As shown in Fig.~\ref{fig:app_efficiency}, the compared methods exhibit
substantially different accuracy--resource trade-offs. MLP has the lowest
runtime and a small memory footprint, but its predictive accuracy remains
below that of the proposed variants. In contrast, \method-F achieves
higher accuracy while retaining relatively low downstream cost,
demonstrating that the pretrained inference backbone can be directly
reused without task-specific optimization.

\method-A achieves the highest accuracy among the compared methods. Its
runtime remains lower than that of several task-specific multi-view
models, although it requires more peak GPU memory than \method-F and
lightweight conventional baselines. Therefore, \method-A should not be
interpreted as the uniformly cheapest method. Instead, the two proposed
variants provide complementary operating points: \method-F prioritizes
direct and efficient deployment, whereas \method-A trades a moderate
amount of additional calibration and memory for improved predictive
performance.

Considering accuracy and runtime jointly, MLP, \method-F, and
\method-A form the empirical Pareto frontier in the plotted comparison.
DCCA, DMF, and ECMGD require comparable or longer execution time without
surpassing \method-F in accuracy, while Co-GCN, RCML, TUNED, and ViHMGD
exhibit substantially lower accuracy and/or larger runtime. These results
support the claim that pretraining a reusable multi-view inference
procedure can reduce repeated downstream adaptation cost.

\end{document}